\documentclass[pdflatex,sn-mathphys-num]{sn-jnl}

\usepackage{graphicx}%
\usepackage{multirow}%
\usepackage{amsmath,amssymb,amsfonts}%
\usepackage{amsthm}%
\usepackage{mathrsfs}%
\usepackage[title]{appendix}%
\usepackage{xcolor}%
\usepackage{textcomp}%
\usepackage{manyfoot}%
\usepackage{booktabs}%
\usepackage{algorithm}%
\usepackage{algorithmicx}%
\usepackage{algpseudocode}%
\usepackage{listings}%
\usepackage{tabulary}
\usepackage{float}
\usepackage{arydshln}
\usepackage{array}

\theoremstyle{thmstyleone}%
\theoremstyle{thmstyletwo}%

\theoremstyle{thmstylethree}%

\begin{document}

\title[Prediction of mortality and resource utilization in critical care: a deep learning approach using multimodal electronic health records with natural language processing techniques]{Prediction of mortality and resource utilization in critical care: a deep learning approach using multimodal electronic health records with natural language processing techniques}


\author[1,2]{\fnm{Yucheng} \sur{Ruan}}\email{yuchengruan@u.nus.edu}
\equalcont{Equal contributions.}
\author[1,2]{\fnm{Xiang} \sur{Lan}}\email{ephlanx@nus.edu.sg}
\equalcont{Equal contributions.}

\author[2]{\fnm{Daniel J.} \sur{Tan}}\email{djtan@u.nus.edu}
\author[3]{\fnm{Hairil Rizal} \sur{Abdullah}}\email{hairil.rizal.abdullah@singhealth.com.sg}
\coseniorauth{Co-senior authors.}
\author*[1,2]{\fnm{Mengling} \sur{Feng}}\email{ephfm@nus.edu.sg}
\coseniorauth{Co-senior authors.}

\affil[1]{\orgdiv{Saw Swee Hock School of Public Health}, \orgname{National University of Singapore}, \orgaddress{\country{Singapore}}}

\affil[2]{\orgdiv{Institute of Data Science}, \orgname{National University of Singapore}, \orgaddress{\country{Singapore}}}

\affil[3]{\orgdiv{Department of Anaesthesiology}, \orgname{Singapore General Hospital}, \orgaddress{\country{Singapore}}}


\abstract{
\textbf{Background}
Predicting mortality and resource utilization from electronic health records (EHRs) is challenging yet crucial for optimizing patient outcomes and managing costs in intensive care unit (ICU). Existing approaches predominantly focus on structured EHRs, often ignoring the valuable clinical insights in free-text notes. Additionally, the potential of textual information within structured data is not fully leveraged. This study aimed to introduce and assess a deep learning framework using natural language processing techniques that integrates multimodal EHRs to predict mortality and resource utilization in critical care settings.

\vspace{5px}

\textbf{Methods}
Utilizing two real-world EHR datasets, we developed and evaluated our model on three clinical tasks (mortality, length of stay (LOS), and surgical duration) with leading existing methods. We also performed an ablation study on three key components in our framework: medical prompts, free-texts, and pre-trained sentence encoder. Furthermore, we assessed the model's robustness against the  corruption in structured EHRs.

\vspace{5px}

\textbf{Results}
Our experiments on two real-world datasets across three clinical tasks showed that our proposed model improved performance metrics by 1.6\%/0.8\% on BACC/AUROC for mortality prediction, 0.5\%/2.2\% on RMSE/MAE for LOS prediction, 10.9\%/11.0\% on RMSE/MAE for surgical duration estimation
compared to the best existing methods. 
It consistently demonstrated superior performance compared to other baselines across three tasks at different corruption rates.

\vspace{5px}

\textbf{Conclusions}
The proposed framework is an effective and accurate deep learning approach for predicting mortality and resource utilization in critical care. The study also highlights the success of using prompt learning with a transformer encoder in analyzing multimodal EHRs. Importantly, the model showed strong resilience to data corruption within structured data, especially at high corruption levels.

\vspace{5px}


}

\maketitle

\section{Background}

Accurate prediction of both mortality and resource utilization are essential for optimizing patient outcomes and supporting cost management in critical care by anticipating the length of stays and resource needs, aiding in budgeting and reducing unnecessary expenditures \cite{iwase2022prediction,saadatmand2023using}.

In recent years, electronic health records (EHRs) have rapidly expanded in medical institutions, which includes diverse patient details, clinical observations, and diagnostic outcomes \cite{jha2009use}. 
Analyzing the EHRs can reveal important patterns and insights that are useful for healthcare providers, researchers, and policymakers.  

Advancements in machine learning have significantly boosted the development of prediction algorithms in critical care, particularly in predicting the mortality and optimizing the resource utilization. Tree-based approaches have demonstrated superior predictive capabilities \cite{nistal2022developing,gao2022prediction}. However, the predominant focus of these approaches has been on structured EHR data (e.g. categorical, numerical, and binary data types). 
With the rise of deep learning, researchers have developed a wide range of frameworks \cite{iwase2022prediction,chen2023deep,kim2025transformer}, while still primarily focusing on structured data. 
Despite of the prevalent structured EHRs, unstructured clinical free-text data, such as clinical notes, diagnostic tests, and preoperative diagnoses, contains valuable clinical information, which can enhance modeling performance using natural language processing (NLP) techniques \cite{zhu2022using}.

In addition, there is an underutilization of textual information within structured data in existing research. For example, categorical features (e.g. diagnosis, prescribed medication) often involve textual descriptions of a patient's condition or treatment context. These descriptions are usually encoded numerically during modeling, ignoring the semantic meanings or hierarchical relationships between different categories.
Another issue with structured EHRs involves data entry errors and data corruption, often due to inconsistent practices and poor management \cite{maletzky2022lifting}. In real clinical situations, syncing structured clinical data between different EHR systems with human intervention can lead to data corruption. With corrupted data, models are prone to be biased, thereby compromising their capability to make consistent and accurate prediction \cite{birman2005accuracy, hersh2013caveats}. Clinical free-texts, on the other hand, can serve as a compensatory source for structured data to mitigate the impact of data corruption \cite{ford2016extracting}.

Our aim in this study \footnote{This study is extended on the short paper accepted by the 23rd International Conference on AI in Medicine (Preprint available at \href{https://arxiv.org/pdf/2303.17408}{https://arxiv.org/pdf/2303.17408})} was to propose and evaluate a deep learning framework for analyzing multimodal EHRs using NLP techniques to predict patient mortality and optimize resource utilization in critical care. Furthermore, we investigated the models' robustness to data corruption in structured EHRs.

\begin{table}[]
\begin{tabular}{p{4cm}p{8cm}}
\hline
\multicolumn{2}{l}{Statement of significance}             \\ \hline
Problem                                                & 
Accurately predicting patient mortality and resource utilization in ICU is important for both patient care and hospital cost management. Current models mostly use structured EHRs while ignoring important information in unstructured clinical free-text. 
\\ \hline
What is Already Known                                  &  
Tree-based machine learning and deep learning methods can effectively predict clinical outcomes using structured EHR data. However, these models often overlook the semantic information in both structured EHRs and unstructured clinical free-texts. Also, data corruption in structured EHRs can bias predictions. 
\\ \hline
What this Paper Adds                                   &  
This paper proposes and evaluates a deep learning framework that leverages both structured EHRs and unstructured free-texts using NLP techniques for predicting mortality and resource utilization in critical care. It demonstrates better predictive performance and robustness to data corruption within structured EHRs.
\\ \hline
Who would benefit from the new knowledge in this paper & 
Clinical informaticians who would like to explore the application of deep learning models and NLP techniques for modeling multimodal EHRs.
\\ \hline
\end{tabular}
\end{table}

\section{Methods}\label{sec3}

\subsection{Study design}

Our framework has three critical stages: cell embedding generation, modality fusion and prediction. We evaluated the frameworks against leading baselines on two real-world datasets: a US dataset called MIMIC-III (Medical Information Mart for Intensive Care III) and a Asian dataset called PASA (Perioperative Anaesthesia Subject Area). 
We also analyzed the model's predictive performance with different levels of corruption in structured EHRs across different tasks.

\subsection{MIMIC-III dataset}
MIMIC-III is a large, publicly available database containing de-identified health records from patients in critical care units at Beth Israel Deaconess Medical Center from US between 2001 and 2012 \cite{goldberger2000physiobank, johnson2016mimic}. We collected structured EHR data and free-text notes from the database. This study included all adult patients aged 18 years or older who were admitted into ICU wards. Patients were excluded if they had missing length of stay or mortality data. We considered only the first record for patients with multiple ICU stays. 

\subsubsection{Structured data}
The patient-level structured data used in MIMIC-III dataset included demographics, vital signs, laboratory tests, medical treatments, and underlying comorbidities. Demographic data at the time of admission were included. 
For vital signs and laboratory tests, the first values documented within the first 24 hours post-admission were used in this study.
All vital sign and laboratory test features were numerical.
Medical treatments, such as sedatives and antibiotics, were included and coded as binary variables to indicate whether the patient received the treatment while comorbidities, such as hypertension and diabetes, were also represented as binary variables. 
Any variables with a missingness rate of 50\% or more were excluded.

In this study, missing data were treated by imputing the mean for continuous features and the mode for categorical features. Data standardization was not required for both categorical and numerical variables.

\subsubsection{Free-text notes}
Free-text notes offer a wealth of clinical information about patients in addition to structured data.
Consequently, NLP techniques, particularly pre-trained language models, can be utilized on these notes to derive deeper clinical insights.
In this study, we selected the longest notes from the first 24 hours of documentation for each admission, since shorter notes are often included in longer ones.
We focused on three types of clinical notes: nursing notes, physician notes, and radiology notes, as these represent the majority of clinical documentation in the MIMIC-III database \cite{zhang2020combining}
Specifically, our study includes two types of nursing notes: \textit{Nursing} and \textit{Nursing/Other}. \textit{Nursing} notes tend to be longer, whereas \textit{Nursing/Other} notes are more frequent \cite{qin2021improving}.

For any missing data in the free-text notes, we imputed it with the sentence "There is no data available."

\subsubsection{Outcomes: mortality and length of stay}
In the MIMIC-III dataset, there are two primary prediction outcomes: mortality and length of stay in the ICU.

Mortality is a crucial outcome for assessing the quality of ICU care. Accurate predictions of patient's mortality can help to evaluate the severity of illness and prioritize care for high-risk patients. 
This, in turn, can improve treatment quality and optimize resource allocation. This prediction task is a binary classification problem, referred as \textbf{MIMIC(Mortality)}.

Length of stay (LOS) refers to the duration that a patient spends in the ICU from admission to discharge. Longer stays are often associated with severe illnesses, higher medical costs and resource utilization. 
Accurately predicting LOS in ICU can not only enhance the quality of critical care but also improve the efficiency of resource allocation. In this study, we term this task as \textbf{MIMIC(LOS)}.

\subsection{PASA dataset}
PASA dataset was retrospectively extracted from Perioperative Anaesthesia Subject Area (PASA) database at Singapore General Hospital (SGD) between 2016 to 2020 \cite{chiew2020utilizing, abdullah2024singhealth}. This study included the preoperative data (structured data and free-text notes) from all patients aged 18 years and above who underwent surgery under anesthesia care at SGH. 
The database received approval from the SingHealth Centralised Institutional Review Board (Singhealth CIRB 2014/651/D, 2020/2915, and 2021/2547), with patient consent being waived.

\subsubsection{Structured data}
The structured data in patient's preoperative anesthesia assessment included demographics, comorbidities, medications, vital signs, laboratory tests, and surgical details.
Demographic information, such as age, height, BMI, were recorded at the time of surgery. 
Comorbidities were documented according to the Revised Cardiac Risk Index \cite{lee1999derivation}, which includes ischemic heart disease, congestive heart failure, diabetes mellitus, cerebrovascular accident, and ASA class.
Administered medications were recorded from the time of patient admission until the start of surgery.
Preoperative vital signs and laboratory tests such as heart rate, prothrombin time, and serum creatinine were the most recent records within 90 days prior to surgery.
Surgery details covered surgical risk, surgical specialty, among other factors. Any variables with a missingness rate of 50\% or more, as well as repetitive variables, were excluded.

For missing data imputation, we followed the procedures in MIMIC-III dataset. Similarly, data standardization was not necessary.

\subsubsection{Free-text notes}
For free-text notes in this dataset, we used four types of patient notes: radiology results, proposed operation, cardio results, and preoperative diagnosis. These notes are the most frequently recorded in the database and offer comprehensive clinical insights into the patient, complementing the structured data.

\subsubsection{Outcome: surgical duration}
Surgical duration refers to the total time taken for a surgical procedure, starting from the initial incision to the completion of the operation. Studies consistently show that longer surgeries are associated with a higher incidence of complications and an increased risk of mortality \cite{cheng2018prolonged, silver2024association}. 
In addition, prolonged surgeries may necessitate more intensive monitoring and support, potentially requiring admission to a critical care unit (ICU) for closer observation. As such, accurate predictions of surgical duration are crucial for effectively coordinating postoperative care resources in critical care settings. We term this task as \textbf{PASA(SD)}.

\subsection{The proposed framework}

In this section, we provide comprehensive details about our proposed framework. To improve the clarity and readability, we prepare a list of essential notations in Table \ref{tab: notation}.

\begin{table}[!h]
\centering
\caption{Some notations and explanations used in the manuscript.}
\begin{tabular}{ll}
\hline
Notations & Explanation \\ \hline
     $m$     &     the number of features in the EHR dataset        \\
      $c_{i,j}$    &    cell value of $i$-th training sample under $j$-th column         \\
       $t_j$   &       medical language template for $j$-th feature       \\ 
      $s_{i,j}$ & medical prompt of $i$-th training sample under $j$-th column\\
      $w_{i,j}^k$ & $k$-th token in the medical prompt $s_{i,j}$\\
      $b_{i,j}^k$ & $k$-th output token embedding in the medical prompt $s_{i,j}$ \\
      $z_{i,j}$ & cell embedding of $i$-th training sample under $j$-th column\\
      $e_{[CLS]}$ & [CLS] token embedding\\
      $u_i^l, r_i^l$ & intermediate embeddings in the block $l$ in transformer encoder\\
      $o_i^l$ & output embeddings of the block $l$ in transformer encoder \\
        $L$ & the number of blocks (layers) in transformer encoder\\
       \hline
\end{tabular}
\label{tab: notation}
\end{table}

Formally, we denote $x_i \in X$ as the $i$-th training sample and $y_i$ as the corresponding labels. Given training dataset $\mathbb{D} = \{x_i, y_i\}_{i=1}^N$ with $N$ examples, the objective of prediction task is to look for a mapping rule $h : \mathcal{X} \rightarrow \mathcal{Y}$ such that a task specific loss function $\mathcal{L}(h)$ is minimized. 

\begin{figure*}[!h]
\centering
\begin{center}
    \includegraphics[scale = 0.4]{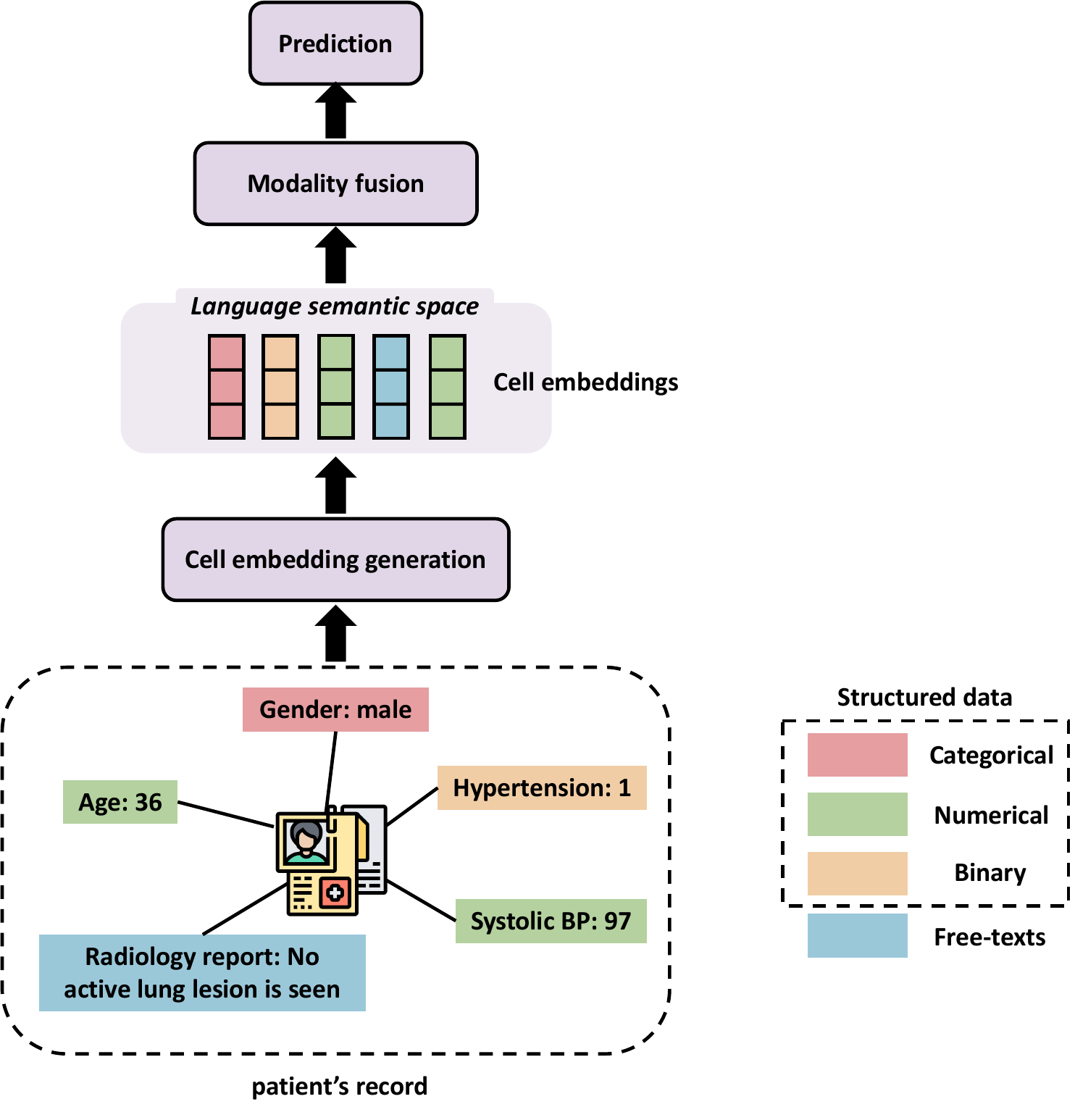}
\end{center}

\caption{The overview of our proposed model. Main stages of the proposed framework: (1) cell embedding generation, (2) modality fusion, (3) prediction. The raw data, including categorical, numerical, binary data in structured EHRs, and free-text data, are first passed through cell embedding generation to produce contextualized cell embeddings in language semantic space. These cell embeddings are then used to generate patient embeddings through transformer encoder in modality fusion for different prediction tasks.}

  \label{fig:framework}
\end{figure*}

A key challenge in learning informative patient embeddings from multimodal EHRs is effectively generating cell embeddings and fusing information from both structured and unstructured data, particularly focusing on the underutilized textual information.

To address this issue, we start by generating cell embeddings using prompting techniques and pre-trained language model in NLP. Subsequently, we perform the modality fusion using a Transformer encoder to generate the patient embedding for final prediction.
The architecture overview is depicted in Figure \ref{fig:framework}.

\subsection{Cell embedding generation}

\begin{figure*}[!h]
\centering
\begin{center}
    \includegraphics[scale = 0.26]{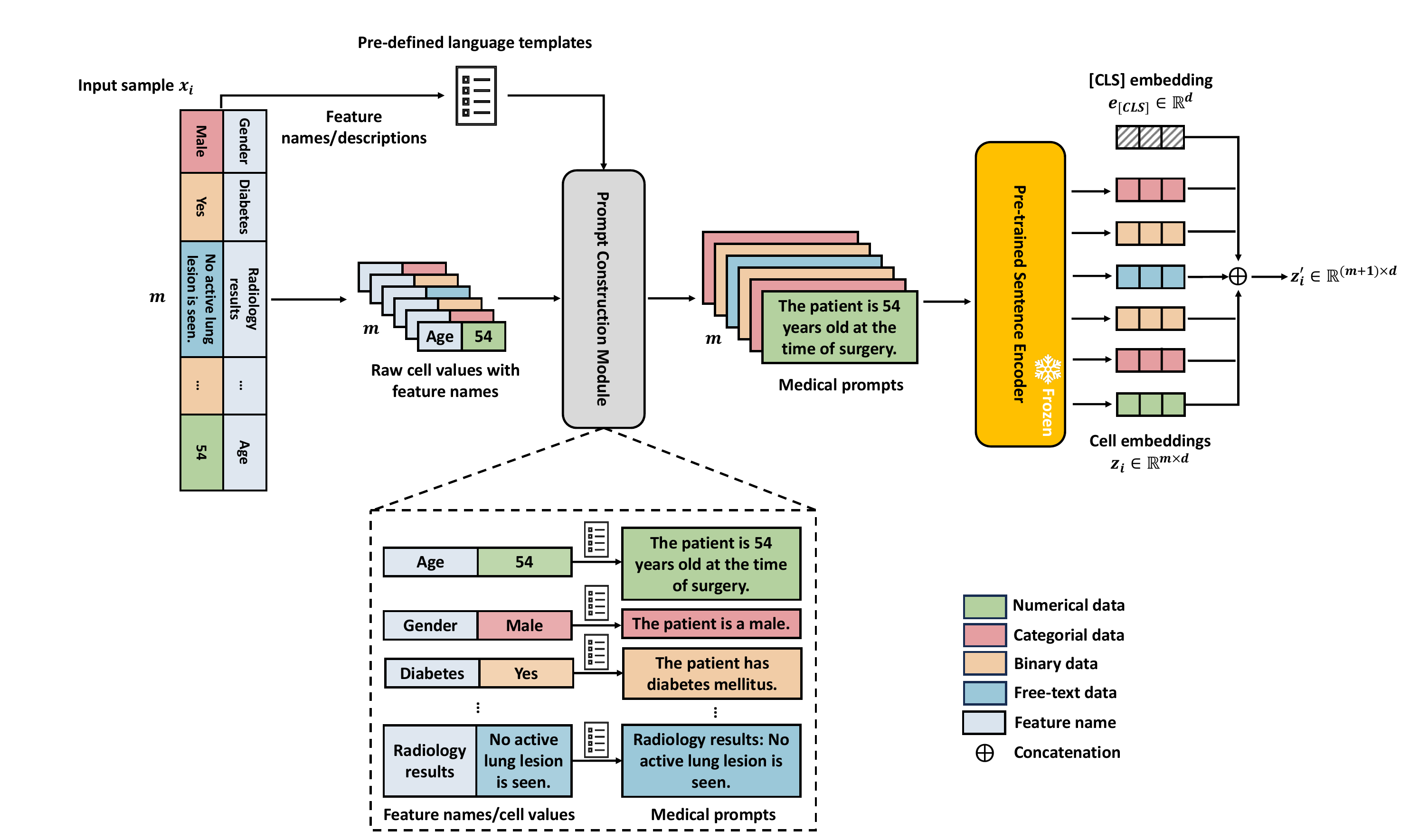}
\end{center}

\caption{Overview of cell embedding generation.}
  \label{fig:tabular_cell_emb}
\end{figure*}

The overview of cell embedding generation is illustrated in Figure \ref{fig:tabular_cell_emb}. It has two critical components: prompt construction module and pre-trained sentence encoder.

\subsubsection{Prompt construction module}

Inspired by the concept of prompt learning, we develop a prompt construction module with medical language templates, which allows the pre-trained language model to more effectively extract contextual information in EHRs, resulting in more informative cell embeddings.
Our language templates convert raw cell values in EHR data into complete natural sentences. These sentences are then processed by the pre-trained sentence encoder to generate sentence embeddings that serve as cell representations. As a result, the contextual information in the raw cell data is also captured.

Let us denote $i$-th training sample in EHR dataset as $x_i = (c_{i,1}, c_{i,2}, \dots, c_{i,m})$. Here, $c_{i,j}$ referes to the cell value as string of $i$-th training sample under $j$-th column, with $m$ indicating the total number of features (columns) in the dataset. In prompt construction module, we pre-define a set of medical language templates for each feature as $T = \{t_1, t_2, \dots, t_m\}$. In this set, $t_j$ denotes the template for $j$-th feature. Using these templates, we can obtain the medical prompts $s_{i,j}$ from cell $c_{i,j}$ by
$c_{i,j} \xrightarrow{t_j} s_{i,j}.$

\begin{table}[!h]
\centering
\caption{Five real examples through prompt construction module.}
{\renewcommand{\arraystretch}{1.5} 
\begin{tabular}{p{1cm}lp{1.5cm}p{3cm}p{3.5cm}}
\hline
Feature           & Type        & Value                          & Template                                                                                                                           & Medical prompts                                     \\ \hline
Age               & Numerical   & 34                             & The patient is {[}value{]} years old at the time of surgery.                                                                       & The patient is 34 years old at the time of surgery. \\
Gender            & Categorical & female                         & The patient is a {[}value{]}.                                                                                                      & The patient is a female.                            \\
Smoking History   & Binary      & yes                            & The patient [yes: has/no: does not have] smoking history.    & The patient has smoking history.                    \\
Diabetes          & Binary      & no                             & The patient [yes: has/no: does not have] diabetes mellitus. & The patient does not have diabetes mellitus.        \\
Radiology results & Free-text   & No active lung lesion is seen. & Radiology results: {[}value{]}                                                                                                      & Radiology results: No active lung lesion is seen.   \\ \hline
\end{tabular}}
\label{tab: template_eg}
\end{table}

For example, assuming our $k$-th feature in our dataset is $weight$, the language template $t_k$ is constructed as "The weight of patient is $c_{i,k}$ kilograms". 
Additional examples are provided in Table \ref{tab: template_eg} to showcase how to generate medical prompts in prompt construction module. Different types of features correspond to different prompt templates. Generally, after prompt construction module, medical prompts $q_i$ of $i$-th training sample, represented as $q_i = (s_{i,1}, s_{i,2}, \dots, s_{i,m})$, are generated for all cells to produce contextualized cell embeddings through the pre-trained sentence encoder.

\subsubsection{Pre-trained sentence encoder}

To further explore the textual information in EHRs, we utilize supervised SimCSE \cite{gao-etal-2021-simcse} from Huggingface to generate harmonized cell embeddings. It's a RoBERTa-based \cite{liu2019roberta} framework for sentence representation generation.

Let us assume the sequence $s_{i,j} = (w_{i,j}^1, w_{i,j}^2, \dots, w_{i,j}^D)$ as the medical prompt of $i$-th training sample under $j$-th column. Here, $w_{i,j}^k$ denotes the $k$-th token in the sequence, while $D$ is the maximum number of tokens that the pre-trained model can take in. 
In pre-trained sentence encoder, $w_{i,j}^1$ is typically the special token $<$s$>$, marking the beginning of sequence. 
For sequences shorter than $D$, the padding token $<$pad$>$ is appended at the end of the sentence to match the maximum sentence length of $D$. Sequences longer than $D$ are truncated to this length. In our study, the fixed length $D$ is set to 512.
The output embedding $b_{i,j}^k$ of $k$-th token is obtained by:

\begin{equation}
    b_{i,j}^1, b_{i,j}^2, \dots, b_{i,j}^D = \mathrm{RoBERTa_{pre}} (w_{i,j}^1, w_{i,j}^2, \dots, w_{i,j}^D).
\end{equation}

Thereafter, we can obtain the cell embedding $z_{i,j}$ by:

\begin{equation}
    z_{i,j} = \mathrm{pooling}(b_{i,j}^1, b_{i,j}^2, \dots, b_{i,j}^D) \in \mathbb{R}^{d}.
\end{equation}

$\mathrm{RoBERTa_{pre}(\cdot)}$ refers to the pre-trained RoBERTa model, and pooling($\cdot$) is the mean pooling layer after token embeddings. The embedding dimension of the pre-trained sentence encoder is denoted by $d$. 
Consequently, the total $m$ cell embeddings are $z_{i} \in \mathbb{R}^{m \times d}$. Following the previous research \cite{gorishniy2021revisiting, wang2022transtab}, [CLS] embedding $e_{[CLS]} \in \mathbb{R}^{d}$ is concatenated with cell embeddings for patient embedding learning as $z'_i$:
\begin{equation}
    z'_i = e_{[CLS]} \oplus z_i \in \mathbb{R}^{(m+1) \times d}.
\end{equation}
As a result, the concatenated embeddings $z'_i$ are obtained for modality exploration.

\subsection{Modality fusion}
After we extract the cell embeddings from both structured EHRs and free-text notes, it's essential to fuse the clinical information from both modalities to enhance predictive accuracy.
To achieve the modality fusion, we adopt the classical transformer encoder \cite{vaswani2017attention,gorishniy2021revisiting} without using positional encoding at the inputs to generate patient embeddings.
The self-attention mechanism within the Transformer architecture can capture the feature relationships regardless of their positions or distances in the EHR dataset.

\begin{figure*}[!h]
\centering
\begin{center}
    \includegraphics[scale = 0.37]{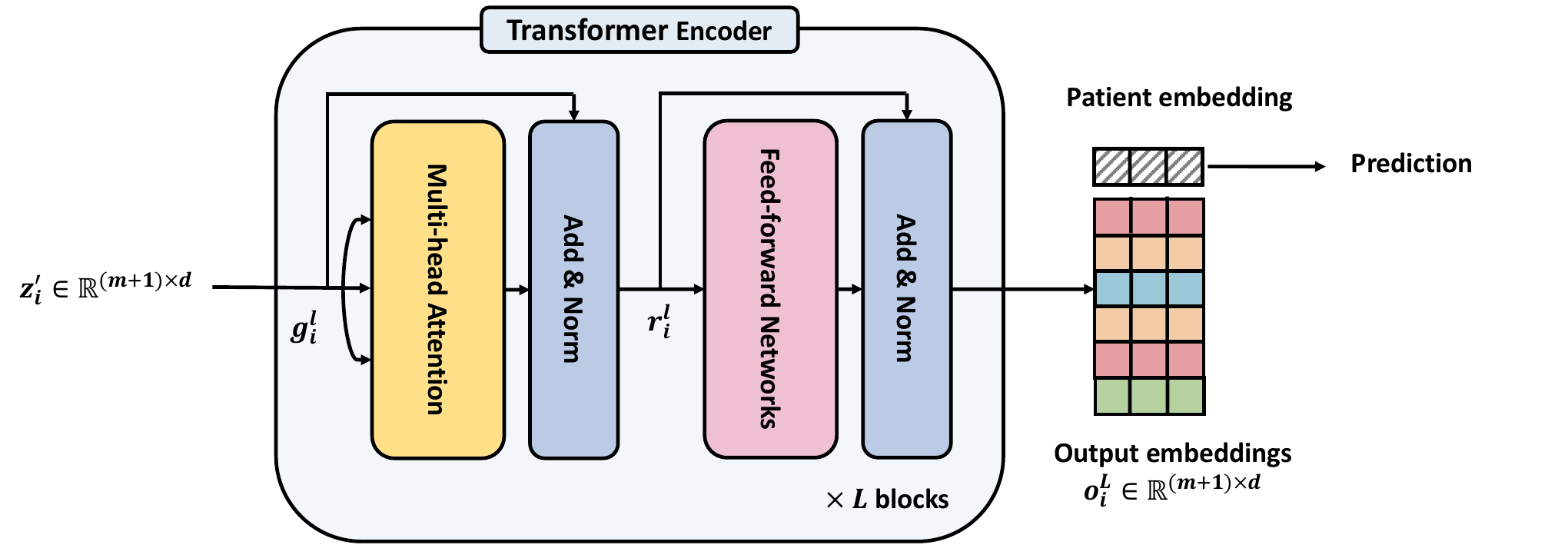}
\end{center}

\caption{Overview of modality fusion with Transformer encoder.}
  \label{fig:tabtran}
\end{figure*}

As illustrated in Figure \ref{fig:tabtran}, the transformer encoder consists of $L$ blocks, and the input embeddings for block $l$ are denoted as $g_i^l$ in $i$-th training sample. Specifically, $g_i^1 = z'_{i} \in \mathbb{R}^{(m+1) \times d}$ at the first block, which are the output embeddings from cell embedding generation. 
The multi-head attention layer is a pivotal component of the transformer architecture. 
It enables the model to focus on different parts of the sequence, thereby contributing to the richer representations. The output embedding $u_i^l \in \mathbb{R}^{(m+1) \times d}$ of multi-head attention layer in block $l$ is defined as follows:
\begin{align}
    u_i^l &= \mathrm{MultiAttention}(Q, K, V) \nonumber, \\
    & = \mathrm{Concat}(head_1, head_2, \cdots, head_H) W_o^l,
\end{align}
where $W_o^l \in \mathbb{R}^{Hd_k \times d}$, and $H$ is the number of attention heads.
The computation for each attention head $head_h$ is defined as:
\begin{align}
    head_h &= \mathrm{Attention}(Q_h, K_h, V_h), \\
    & = \mathrm{softmax}(\frac{Q_h K_h^T}{\sqrt{d_k}}V_h),
\end{align}
where $Q_h = g_i^l W_{h, q}^l$, $K_h = g_i^l W_{h, k}^l$, and $V_h = g_i^l W_{h, v}^l$. The weight matrices $W_{h, q}^l, W_{h, k}^l, W_{h, v}^l \in \mathbb{R}^{d \times d_k}$ are tunable parameters during training. 
After the multi-head attention layer, the output $r_i^l \in \mathbb{R}^{(m+1) \times d}$ of the layer normalization layer is then calculated as below:  
\begin{equation}
    r_i^l = \mathrm{LayerNorm}(u_i^l + g_i^l; \gamma_1, \beta_1),
\end{equation}
where $\gamma_1, \beta_1 \in \mathbb{R}^{d}$ are the parameters that scale and shift the normalized values.

Next, a two-layer feed-forward neural network ($\mathrm{FFN}(x) = \mathrm{max}(0, xW_1 + b_1)W_2 + b_2$) is used to transform the $r_i^l$ to output embeddings $o_i^l \in \mathbb{R}^{(m+1) \times d}$ of the block $l$ in transformer encoder with Add and LayerNorm, which are the input embeddings $g_i^{l+1}$ for block $l+1$. So, 
\begin{equation}
    g_i^{l+1} = o_i^l = \mathrm{LayerNorm}(\mathrm{FFN}(r_i^l) + r_i^l; \gamma_2, \beta_2),
\end{equation}
where $\gamma_2, \beta_2 \in \mathbb{R}^{d}$ are the parameters to scale and shift the normalized values. 

Finally, we take $o_i^L \in \mathbb{R}^{(m+1) \times d}$ as the output embeddings of the entire transformer encoder, which are used for further prediction.

\subsection{Prediction}
Since we have the encoded output embeddings $o_i^L \in \mathbb{R}^{(m+1) \times d}$ from previous modality fusion, we take the [CLS] embedding $e'_{[CLS]} \in \mathbb{R}^d$ as the patient embedding. The patient embedding is used for different medical tasks through a shallow feed-forward network (FFN). In general, the predictive tasks can be categorized into two categories: classification and regression.

\subsubsection{Classification}
For classification problem, a feed-forward neural network (FNN) with one hidden layer is implemented for prediction, and Cross-entropy loss has been used for optimization. Let $y_i = (y_{i,1}, y_{i, 2}, \cdots, y_{i, K})$ be the one-hot encoding of the true label for the $i$-th sample, where $K$ is the number of categories. Then the loss function is defined as follows:
\begin{equation}
    \mathcal{L}_{CE} (y, x) = - \frac{1}{N} \sum_{i=1}^{N} \sum_{k=1}^{K} w_k y_{i, k} \mathrm{log} P(o_{i, k} | x_i),
\end{equation}
\begin{equation}
    P(o_{i, k} | x_i) = \frac{\mathrm{exp}(o_{i, k})}{\sum_{j=1}^{K} \mathrm{exp}(o_{i, j})},
\end{equation}
where $P(o_{i, c} | x_i)$ is the probability of category $k$ in $i$-th training sample, $w_k$ is the class weight, and $o_i = (o_{i,1}, o_{i, 2}, \cdots, o_{i, K})$ is the final outputs of the classification networks from $i$-th training sample.

\subsubsection{Regression}
For regression problem, a FNN with one hidden layer is implemented as well for prediction, and Mean Squared Error (MSE) is used for parameter optimization. Let $y_i$ be the ground-truth label for the $i$-th sample. The loss function is formulated as below:
\begin{equation}
    \mathcal{L}_{MSE} (y, x) = - \frac{1}{N} \sum_{i=1}^{N} (y_i - o_i')^2,
\end{equation}
where $o_i'$ is the final outputs of the regression networks from $i$-th training sample.

\subsection{Baselines}
To better demonstrate the performance of the proposed model, we conducted comparison experiments on the both datasets with the following baselines that achieved great prediction results.

\begin{itemize}
    \item \textbf{Random Forest} \cite{breiman2001random} is a tree-based ensemble learning method, which operates by constructing a multitude of decision trees during training and outputs the class for prediction.
    \item \textbf{XGBoost} \cite{chen2016xgboost} is another tree-based ensemble learning method that aggregates predictions from multiple weak models. Renowned for its efficiency and high predictive performance, it remains a robust and widely used model in clinical prediction \cite{xi2018research,gao2022prediction, nistal2022developing}.
    \item \textbf{MLP} \cite{gorishniy2021revisiting} is a type of artificial neural network designed for supervised learning tasks.
    \item \textbf{ResNet} \cite{gorishniy2021revisiting}, originally developed for computer vision \cite{he2016deep}, has been adapted for tabular structured data modeling, utilizing residual blocks as robust baselines
    \item \textbf{TabNet} \cite{arik2021tabnet} is a groundbreaking transformer-based model designed for tabular prediction. It effectively manages tabular structured data by integrating sequential models and attention mechanisms.
    \item \textbf{TabTransformer} \cite{huang2020tabtransformer} employs transformers with self-attention mechanisms to transform categorical feature embeddings into robust contextual embeddings, resulting in improved prediction accuracy.
    \item \textbf{FT-Transformer} \cite{gorishniy2021revisiting} introduces the Feature Tokenizer, which transforms all features, both categorical and numerical, into embeddings. These embeddings are subsequently input into transformer layers to enable the generation of robust predictions. It has emerged as a state-of-the-art (SOTA) model surpassing tree-based models in tabular prediction.
    \item \textbf{TransTab} \cite{wang2022transtab} integrates word embedding learning to incorporate textual information in the modeling process. This involves designing ad-hoc embedding modules for each modality in both structured data and unstructured texts, coupled with a transformer-based architecture for predictions.
\end{itemize}

\subsection{Implementation details}

The dataset was randomly split into training, validation, and testing sets in a ratio of 3:1:1
In our proposed model, we used the supervised SimCSE \cite{gao-etal-2021-simcse} as the pre-trained sentence encoder. Each word token was mapped into a 768-dimensional embedding, and the entire encoder was frozen in the training process. The transformer encoder consisted of 6 basic transformer encoder layers, each with 6 heads in the attention layer. Throughout training, we used the Adam optimizer \cite{kingma2014adam} to update gradients with a learning rate of 1e-5 across all tasks. A mini-batch size was set to 256, and the maximum number of epochs was set to 100, with early stopping applied. 
Owing to significant GPU memory demands, particularly in our case, where the language encoder needs to process multiple times in one sample, we designed a two-step training scheme. In the scheme, we extracted all cell embeddings first, facilitating the subsequent training of the transformer encoder with these embeddings in the second step. For baselines, we optimized the hyperparameters to establish the reliable baselines, ensuring the feasibility of head-to-head comparisons. 

In classification task MIMIC(Mortality), we applied a straightforward class weighting technique \footnote{https://scikit-learn.org/stable/modules/generated/sklearn.utils.class\_weight.compute\_class\_weight.html} based on relative class frequencies to address data imbalance, given that it is not the primary focus of our study. The ratio of positive weight to negative weight was set at 7.5:1. 
\label{class_weighting}

We repeated model training 5 times with different random seeds and reported the average metrics, ensuring statistically stable results in this study.

\subsection{Evaluation Metrics}
For model evaluation and comparison, we demonstrate the evaluation metrics for classification and regression tasks separately.
\subsubsection{Classification}
In classification task, we reported balanced accuracy (BACC) and the area under the ROC curve (AUROC) because the medical dataset for the classification task in this study (mortality prediction) was imbalanced. These two metrics are better to evaluate the model performance.

\begin{equation}
    \text{Balanced accuracy} = \frac{\text{Sensitivity} + \text{Specificity}}{2},
\end{equation}
\begin{equation}
    \text{Sensitivity} = \frac{\text{TP}}{\text{TP} + \text{FN}},
\end{equation}

\begin{equation}
    \text{Specificity} = \frac{\text{TN}}{\text{FP} + \text{TN}},
\end{equation}
where TP, FP, TN, FN are true positive, false positive, true negative and false negative respectively in classification.
\subsubsection{Regression}
In regression task, we reported the root mean squared error (RMSE) and mean absolute error (MAE). Given $y_i$ as the ground truth of $i$-th data sample and $h(x_i)$ as the predicted label, 
\begin{equation}
    \text{RMSE} = \sqrt{\frac{1}{N} \sum_{i=1}^N (y_i - h(x_i))^2},
\end{equation}
\begin{equation}
    \text{MAE} = \frac{1}{N} \sum_{i=1}^N |y_i - h(x_i)|.
\end{equation}

\begin{table}[!h]
\centering
\caption{Selected patient's characteristics for MIMIC-III dataset. Data are represented as median [Q1-Q3] or number (\%).}
\begin{tabular}{ll}
\hline
Characteristics               & Cohort (n=38468)        \\ \hline
Admission age (years)                & 64.39 {[}51.60-76.20{]} \\
Gender (\%)                   &                         \\
\quad Male                          & 21777 (56.61\%)         \\
\quad Female                        & 16691 (43.39\%)         \\
Ethnicity (\%)                &                         \\
\quad White                         & 27443 (71.34\%)         \\
\quad Black                         & 2948 (7.66\%)           \\
\quad Hispanic                      & 1254 (3.26\%)           \\
\quad Asian                         & 909 (2.36\%)            \\
\quad Others                        & 5914 (15.37\%)          \\
Weights (kg)                    & 79.00 {[}67.30-93.00{]} \\
Hypertension (\%)             & 17981 (46.74\%)         \\
Alcohol abuse (\%)            & 1756 (4.56\%)           \\
Diabetes mellitus (\%)                 & 10356 (26.92\%)         \\
Cerebrovascular accident (\%) & 1448 (3.76\%)           \\
Congestive heart failure (\%) & 10127 (26.33\%)         \\
Ischemic heart disease (\%)   & 13666 (35.53\%)         \\
Mortality (\%)                & 4540 (11.80\%)          \\
Length of stay (days)                & 2.10 {[}1.19-4.09{]}    \\ \hline
\end{tabular}
\label{tab:data_mimic}
\end{table}

\section{Results}

\subsection{Data description}

The selected characteristics of the patients in the study are demonstrated in Tables \ref{tab:data_mimic} and \ref{tab:data_pasa}. Discrete variables are presented as numbers and percentages, while continuous variables are shown as medians with interquartile ranges ([Q1-Q3]).

In the MIMIC-III dataset (Table \ref{tab:data_mimic}), there were 38,468 patients in the cohort, with a median admission age of 64.39 years [51.60-76.20], and a higher proportion of male patients (56.61\%). Most patients were White (71.34\%), followed by Black (7.66\%), Hispanic (3.26\%), and Asian (2.36\%). Nearly half had hypertension (46.74\%), and 4.56\% had alcohol abuse. Common cardiovascular comorbidities included diabetes mellitus (26.92\%), cerebrovascular accident (3.76\%), congestive heart failure (26.33\%), and ischemic heart disease (35.53\%). Clinically significant outcomes included mortality rate, with 4,540 patients (11.80\%) dying during their ICU stay, and a median ICU stay of 2.10 days [1.19-4.09].

The patient cohort in PASA dataset (Table \ref{tab:data_pasa}) included 71,082 patients, with a median age of 59.00 years [46.0-69.0] at surgery time and a balanced gender distribution (males: 49.97\%). The majority were Chinese (73.29\%), followed by Indian (9.96\%), Malay (9.41\%), and Caucasian (0.62\%). General comorbidities included smoking history (10.69\%), alcohol consumption (3.18\%), and hypertension (44.9\%). Cardiovascular comorbidities were diabetes mell1`itus (4.41\%), cerebrovascular accident (3.53\%), congestive heart failure (3.07\%), and ischemic heart disease (12.79\%). The main clinical outcome was the surgical duration, with a median of 1.83 hours [1.07-3.25].

Table \ref{tab:feat_des} indicates that the PASA dataset features more variables in different types for this study compared to the MIMIC-III dataset.

\begin{table}[!h]
\centering
\caption{Selected patient's characteristics for pasa dataset. Data are represented as median [Q1-Q3] or number (\%).}
\begin{tabular}{ll}
\hline
Characteristics                   & Cohort (n=71082)        \\ \hline
Age at time of surgery (years)             & 59.00 {[}46.00-69.00{]} \\
Gender (\%)                       &                         \\
\quad Male                              & 35518 (49.97\%)         \\
\quad Female                            & 35564 (50.03\%)         \\
Race (\%)                         &                         \\
\quad Chinese                           & 52099 (73.29\%)         \\
\quad Indian                            & 7079 (9.96\%)           \\
\quad Malay                             & 6692 (9.41\%)           \\
\quad Caucasian                         & 444 (0.62\%)            \\
\quad Others                            & 4768 (6.71\%)           \\
BMI                               & 24.80 {[}22.1-28.2{]}   \\
Smoking history (\%)                   & 7601 (10.69\%)          \\
Alcohol consumption (\%)          & 2260 (3.18\%)           \\
Hypertension (\%)                 & 28745 (40.44\%)         \\
Diabetes mellitus (\%) & 3134 (4.41\%)           \\
Cerebrovascular accident (\%)     & 2507 (3.53\%)           \\
Congestive heart failure (\%)     & 2182 (3.07\%)           \\
Ischemic heart disease (\%)       & 9091 (12.79\%)          \\
Surgical duration (hours)         & 1.83 {[}1.07-3.25{]}    \\ \hline
\end{tabular}
\label{tab:data_pasa}
\end{table}

\begin{table}[!h]
\centering
\caption{Feature description for MIMIC-III and PASA  datasets.}
\begin{tabular}{lll}
\hline
 Description                        & MIMIC-III & PASA \\ \hline
Number of categorical features & 3    & 12\\
Number of continuous features  & 22    & 26 \\
Number of binary features      & 13    & 32\\
Number of free-text features        & 4     & 4\\

\hline
\end{tabular}
\label{tab:feat_des}
\end{table}

\subsection{Model performance}

\begin{table}[!h]
\centering
\caption{The results of model comparisons on MIMIC-III and PASA datasets for three different tasks. The best results among all models are in bold while the best results within the same groups are underlined.}
\begin{tabular}{lllllll}
\toprule
\multirow{2}{*}{Model}  & \multicolumn{2}{l}{MIMIC(Mortality)} & \multicolumn{2}{l}{MIMIC(LOS)} & \multicolumn{2}{l}{PASA(SD)} \\ \cmidrule{2-7} 
                                             & BACC$\uparrow$             & AUROC$\uparrow$ & RMSE$\downarrow$          & MAE$\downarrow$ & RMSE$\downarrow$        & MAE$\downarrow$            \\ \midrule
\multicolumn{7}{c}{Structured data only} \\
\midrule
Random Forest                           & 0.722            & 0.809   & 5.820         & 3.077   & 1.410        & 0.896      \\
XGBoost                                 & 0.761            & 0.846  & 5.731         & 3.003   & 1.364        & 0.845       \\
MLP                                     & 0.754            & 0.836  & 5.771         & 3.027  & 1.376        & 0.851         \\
ResNet                                  & 0.759            & 0.842  & 5.752         & 2.983   & 1.368        & 0.840       \\
TabNet                                  & 0.748            & 0.826   & 5.846         & 3.081   & 1.423        & 0.876        \\
TabTransformer                          & 0.754            & 0.835  & 5.754         & 3.053  & 1.353        & 0.848        \\
FT-Transformer                          & 0.762            & \underline{0.848}   & \underline{5.722}         & 3.002   & 1.343        & 0.828      \\ 
 \hdashline[2.5pt/5pt] 
Ours (w/o free-texts)    & \underline{0.765}            & 0.844                        & 5.755         & \underline{2.964}           & \underline{1.340}        & \underline{0.825}          \\
\midrule
\multicolumn{7}{c}{Structured data + Free-text notes} \\
\midrule
TransTab                                & 0.741            & 0.827  & 5.801         & 3.100   & 1.349        & 0.860       \\
 \hdashline[2.5pt/5pt] 
Ours                               & \underline{\textbf{0.774}}            & \underline{\textbf{0.855}}  & \underline{\textbf{5.692}}         & \underline{\textbf{2.918}}   & \underline{\textbf{1.197}}        & \underline{\textbf{0.737}}       \\ 
$\mathrm{\Delta}$ over best baseline                 &  1.6\%            & 0.8\%   &  0.5\%         &  2.2\% &
 10.9\%       &  11.0\%         \\
\bottomrule
\multicolumn{7}{l}{$\mathrm{\Delta}$: model performance improvements}
\end{tabular}
\label{tab:model_comparison}
\end{table}

We conducted extensive experiments to compare our model with other baseline approaches, which were grouped into two: models using structured data only and modeling using structured data and free-text notes. The results, presented in Table \ref{tab:model_comparison}, demonstrate the superior performance of our proposed framework across various tasks. Specifically, in the MIMIC(Mortality) task, our proposed model surpassed the best baseline with a BACC/AUROC of 0.774/0.855.  Similarly, in the MIMIC(LOS) task, our model outperformed other baselines, yielding an RMSE/MAE of 5.692/2.918. In the PASA(SD) task, our framework achieved the best performance among the current SOTA models, with an RMSE/MAE of 1.197/0.737.

To provide further insight, our framework exhibited a performance increase of about 1.6\%/0.8\% in BACC/AUROC in the MIMIC(Mortality) task over the best baseline. It also demonstrated performance improvements, reducing RMSE/MAE by approximately 0.5\%/2.2\% in the MIMIC(LOS) task and 10.9\%/11.0\% in the PASA(SD) task compared to the best baseline. 

Furthermore, when assessing models using structured data only, our proposed framework achieved performance on par with the state-of-the-art models.

\subsection{Ablation study}

We conducted an ablation study to assess the effectiveness of three critical modules of our proposed model: the medical prompts, the free-texts, and the pre-trained sentence encoder.

\begin{table}[!h]
\caption{The ablation RMSE/MAE results in PASA and MIMIC-III datasets for three different tasks.}
\begin{tabular}{lp{0.8cm}p{0.8cm}p{0.8cm}p{0.8cm}p{0.8cm}p{0.8cm}}
\toprule
\multirow{2}{*}{Model}           & \multicolumn{2}{l}{MIMIC(Mortality)} & \multicolumn{2}{l}{MIMIC(LOS)} & \multicolumn{2}{l}{PASA(SD)} \\ \cmidrule{2-7} 

                                 &  BACC$\uparrow$            & AUROC$\uparrow$
                                 & RMSE$\downarrow$         & MAE$\downarrow$          & RMSE$\downarrow$        & MAE$\downarrow$                    \\ \midrule
Complete model          & 0.774            & 0.855             & 5.692         & 2.918            & 1.197        & 0.737         \\
\quad w/o medical prompts    & 0.745            & 0.822                  & 5.817         & 3.061           & 1.263        & 0.777          \\
\quad w/o free-texts    & 0.765            & 0.844                        & 5.755         & 2.964           & 1.340        & 0.825          \\
\quad w/o pre-trained sentence encoder   & 0.504            & 0.509      & 6.117         & 3.109            & 2.509        & 1.863         \\ \bottomrule
\end{tabular}

\label{tab: ablation}
\end{table}

\subsubsection{Medical prompts}
In the prompt construction module, medical language templates are devised to enable a pre-trained language model to generate comprehensive cell representations by converting medical data in cells into natural language sentences. To assess the impact of medical prompts on predictive performance, experiments were conducted whereby the pre-trained sentence encoder was trained on the raw cell values rather than the medical prompts. 
The experimental results in Table \ref{tab: ablation} reveal that the model incorporating medical prompts resulted in an increase in BACC/AUROC by 3.9\%/4.0\% for MIMIC(Mortality) task.
The model with medical prompts exhibited a reduction of 2.1\%/4.7\% for the MIMIC(LOS) task and 5.2\%/5.1\% in RMSE/MAE for the PASA(SD) task.

\subsubsection{Free-texts}
Since our proposed framework takes the free-text information into account for modeling, we conducted a comparative analysis to assess the significance of free-texts in EHRs by examining the performance of the models with and without free-text columns. 
As shown in Table \ref{tab: ablation}, the model with free-texts achieved better predictive performance with an increase in BACC/AUROC by 1.2\%/1.3\% in MIMIC(Mortality) task. In the MIMIC(LOS) and PASA(SD) tasks, the model incorporating additional free-text information also demonstrated enhanced performance, with a drop of 1.1\%/1.6\% and 10.7\%/10.7\% in RMSE/MAE respectively.

\subsubsection{Pre-trained sentence encoder}
To study the effect of the pre-trained sentence encoder in our model, we replaced it with a vanilla encoder (with the same architecture but random initialization) for performance comparison. 
The results in Table \ref{tab: ablation} show that the model with the pre-trained sentence encoder significantly increased the BACC/AUROC by 53.5\%/68.0\% in MIMIC-III(Mortality) task. It also reduced the RMSE/MAE by 6.9\%/6.1\% in the MIMIC-III(LOS) task and 52.3\%/60.4\% in the PASA(SD) task.

\begin{figure}[!h]
\centering
\begin{center}
    \includegraphics[scale = 0.45]{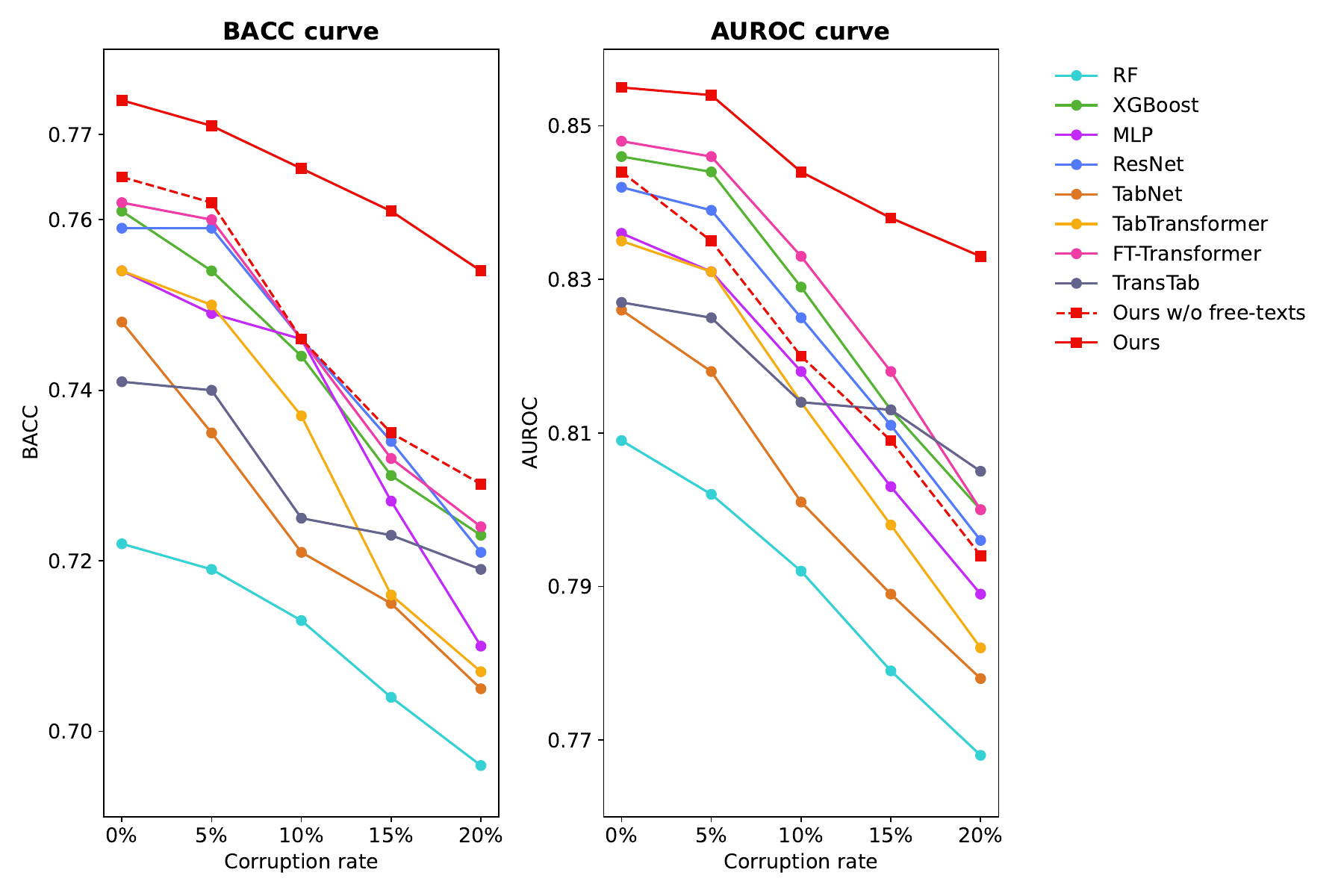}
\end{center}

\caption{Test results with different data corruption rates in MIMIC-III(Mortality) task. Higher metrics indicate better model performance}
  \label{fig:mimic_mortality}
\end{figure}

\begin{figure}[!h]
\centering
\begin{center}
    \includegraphics[scale = 0.45]{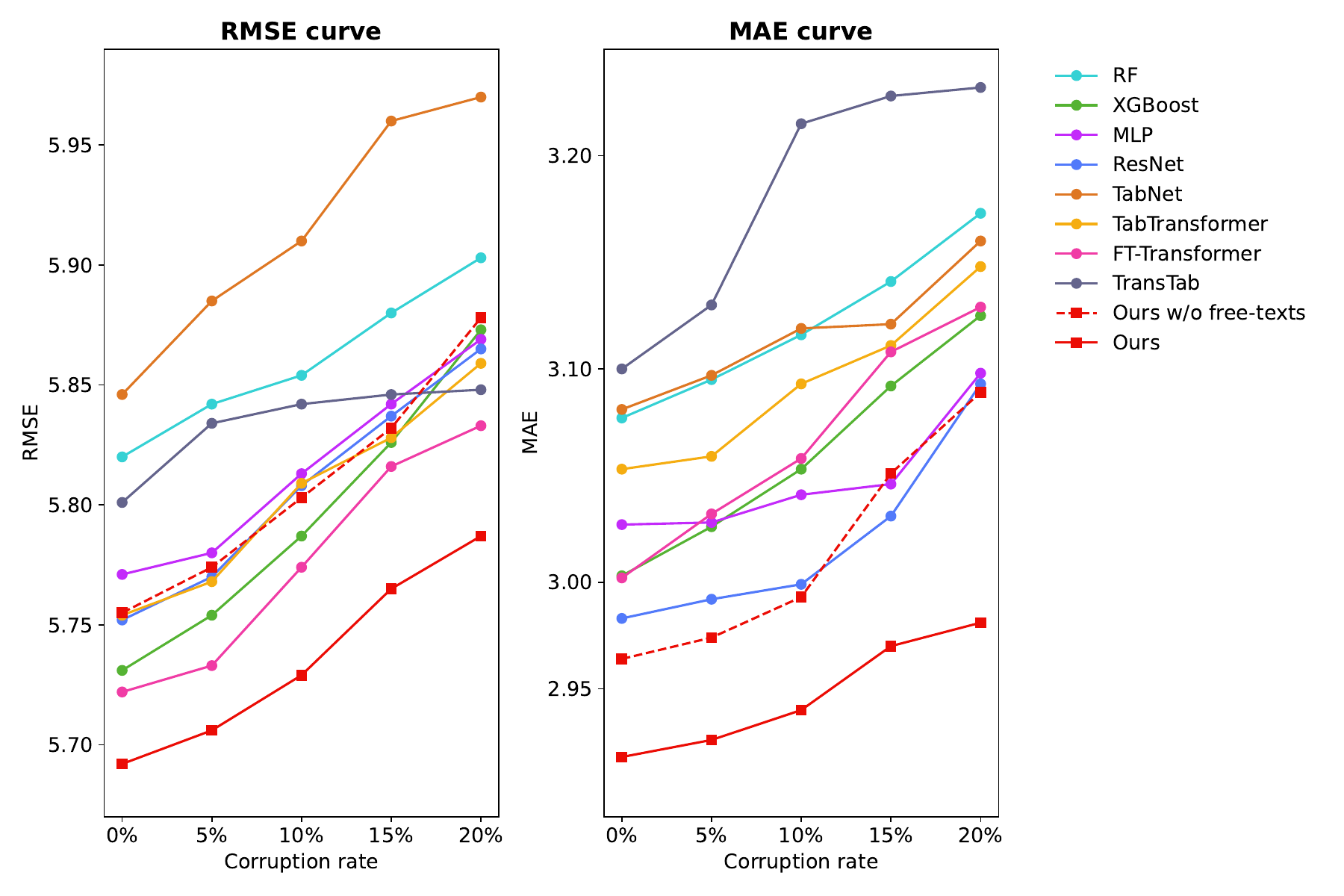}
\end{center}

\caption{Test results with different data corruption rates in MIMIC-III(LOS) task. Lower metrics indicate better model performance}
  \label{fig:mimic_los}
\end{figure}

\begin{figure}[!h]
\centering
\begin{center}
    \includegraphics[scale = 0.45]{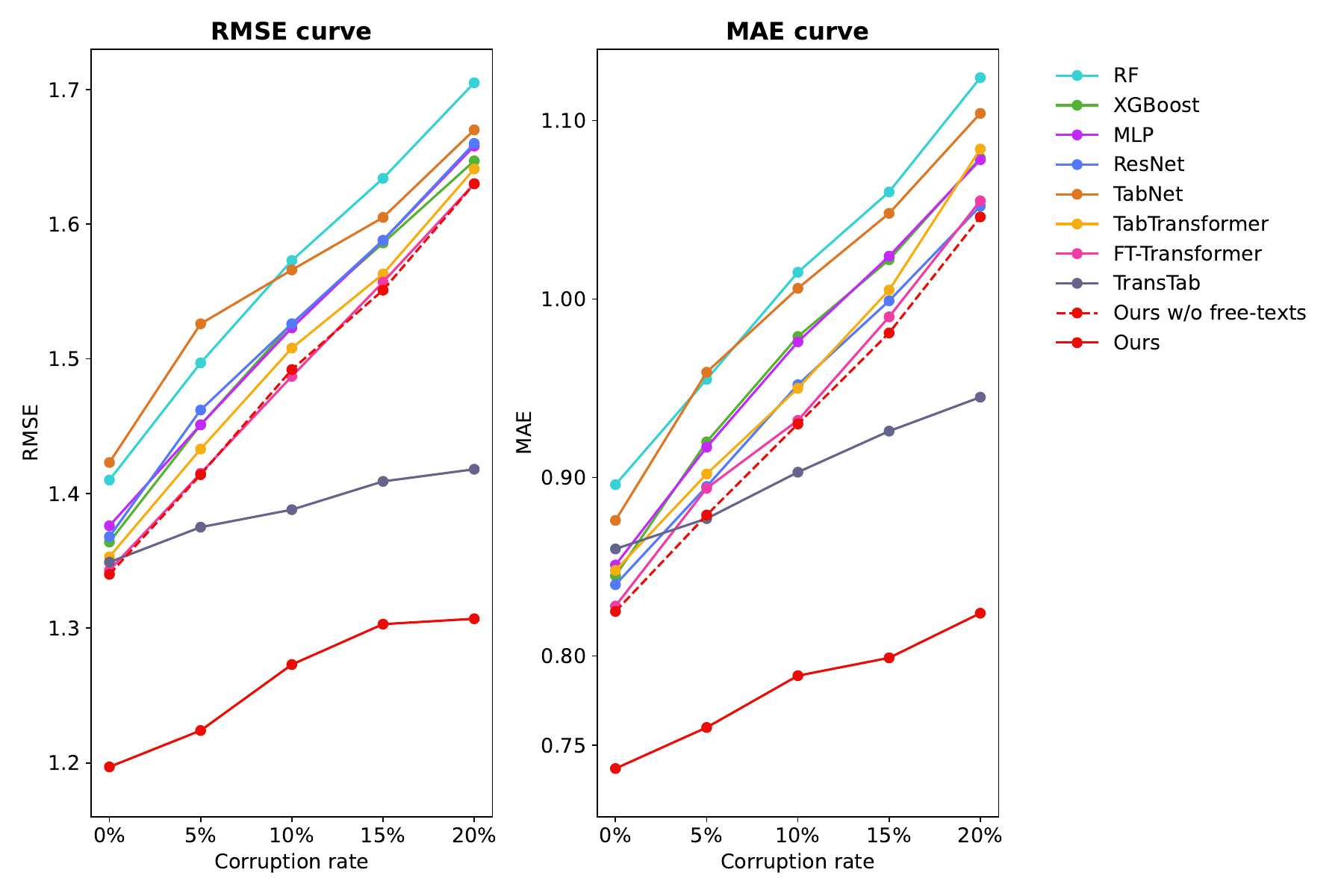}
\end{center}

\caption{Test results with different data corruption rates in PASA(SD) task. Lower metrics indicate better model performance.}
  \label{fig:pasa_sd}
\end{figure}

\subsection{Robustness Study to Data Corruption}
In order to evaluate the robustness of our proposed model in real-life scenarios and validate its resilience to corruption in structured EHRs, we conducted the experiments by only corrupting the structured data in the datasets at the following rates: 5\%, 10\%, 15\% and 20\%. We implemented random feature corruption following the approach described in \cite{bahri2021scarf}. The experimental results on three tasks vs corruption rates are shown in Figures \ref{fig:mimic_mortality}, \ref{fig:mimic_los}, and \ref{fig:pasa_sd}.
Our proposed model consistently demonstrated superior performance compared to other baseline models across three tasks at different corruption rates. 

\section{Discussion}

This is the first study to propose and evaluate a deep learning framework with prompt-based NLP techniques using structured EHRs and free-text notes for predicting patient mortality and optimizing resource utilization in critical care. The main benefit of our model is the capability to analyze multimodal EHRs by projecting all modalities into a harmonized language semantic space through prompt learning and a pre-trained language model, allowing the transformer encoder to produce robust predictions. 
Additionally, our experiments indicate that our model achieved better performance compared to other baselines to data corruption in structured EHRs.

\subsection{Model comparisons}
The consistent superiority of our model across three tasks underscores its effectiveness. We observed variations in model performance between datasets. The PASA dataset exhibited significantly enhanced model performance compared to the MIMIC-III dataset, possibly due to the notably lower frequency of missing clinical free-texts in the PASA dataset. Specifically, there were about 15\% missing values in PASA dataset, compared to about 65\% in MIMIC data. 
We also evaluated our framework using structured data only, and our model remained effective against leading models, indicating that the prompting techniques can effectively utilize the textual information in structured data.
These observations emphasize the crucial role of textual information in both structured data and free-text notes in multimodal learning.

Existing machine learning approaches have been extensively investigated in mortality prediction and resource optimization, with tree-based models achieving the best results \cite{iwase2022prediction,nistal2022developing,gao2022prediction,riahi2023improving}.
Our study confirms similar findings, showing that XGBoost maintained strong predictive performance in modeling structured EHRs.
Our results also align with the study \cite{gorishniy2021revisiting}, which highlighted FT-Transformer as a compelling competitor to XGBoost in existing transformer-based deep learning approaches.

Importantly, it is worth noting that, although TransTab incorporated textual information through the learning of word embeddings, these embeddings were not adequately learned with the constraints of relatively limited medical data when compared to the extensive data used in training large language models. As a consequence, this inadequacy led to a degradation in predictive performance when compared to state-of-the-art transformer-based models.

\subsection{Data imbalance in clinical practice}

Data imbalance significantly complicates classification problems, particularly when the minority class is more crucial. Models trained on imbalanced data often fail to effectively identify minority instances, which can lead to severe consequences \cite{sowjanya2023effective, zhu2024processing}, especially in clinical tasks, such as cancer prediction \cite{fotouhi2019comprehensive} and cardiovascular disease prediction \cite{drouard2024exploring}.

\begin{figure}[!h]
\centering
\begin{center}
    \includegraphics[scale = 0.5]{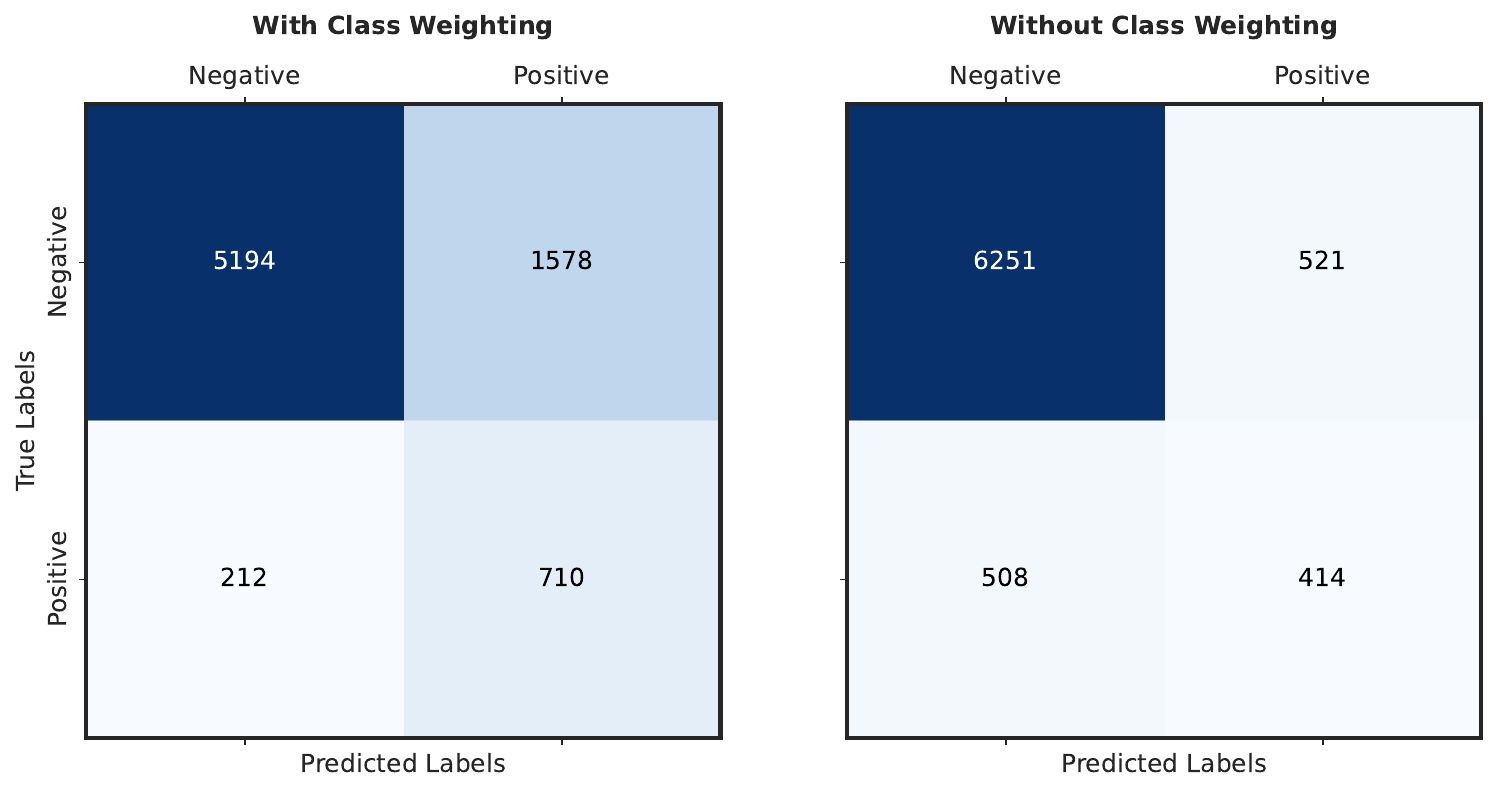}
\end{center}

\caption{Confusion matrices of models with and without class weighting (threshold=0.5) on MIMIC-III(Mortality) task.}
  \label{fig:confusion_matrix}
\end{figure}

\begin{figure}[!h]
\centering
\begin{center}
    \includegraphics[scale = 0.35]{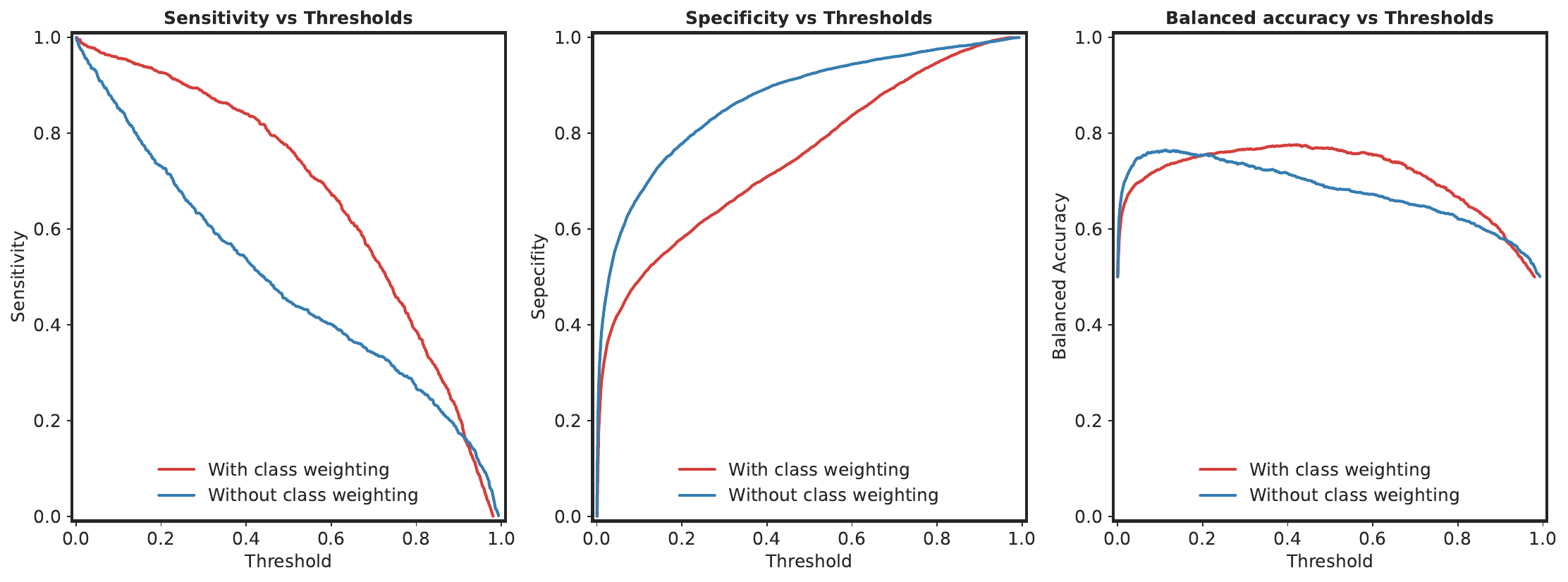}
\end{center}

\caption{Sensitivity, specificity and balanced accuracy results of models with and without class weighting on different thresholds on MIMIC-III(Mortality) task.}
  \label{fig:class_weighting_threshold}
\end{figure}

We evaluated the impacts of data imbalance in the MIMIC(Mortality) task by training our models with and without a class weighting technique, as described in Section \ref{class_weighting}. Figure \ref{fig:confusion_matrix} indicates that applying class weighting can help address data imbalance in mortality prediction, potentially reducing false negatives. It suggests that more critically ill patients, who are at high risk of dying during their ICU stay, can be identified for early intervention. 
Furthermore, Figure \ref{fig:class_weighting_threshold} demonstrates that models using class weighting generally achieved higher sensitivity across different thresholds compared to those without.
However, there is a trade-off between sensitivity and specificity; higher sensitivity leads to lower specificity, resulting in more false positives and potential wastage of medical resources. In this case, balanced accuracy offers a comprehensive evaluation of both sensitivity and specificity. As depicted in the balanced accuracy versus thresholds curves (Figure \ref{fig:class_weighting_threshold}), models with class weighting generally outperformed those without in balanced accuracy across most thresholds. In real practice, clinical professionals can adjust the threshold to align with clinical objectives.

\subsection{Ablation study}
All three components (medical prompts, free-text notes and pre-trained sentence encoder) have been proved to make substantial contributions in enhancing the model predictive capability. Notably, the model with a pre-trained sentence encoder yielded the most substantial performance gains, highlighting the indispensability of this component in effectively extracting cell embeddings.

\subsection{Robustness study to data corruption}
The metric curves of our model generally showed a slower increase compared to baseline models as the corruption rate rose, indicating a significant level of resilience, particularly when structured EHRs is highly corrupted. In contrast, our model without free-texts behaved similarly to the baseline models. The difference in corruption curves is more pronounced in the PASA dataset than in the MIMIC-III dataset, possibly due to fewer missing values in the free-texts of the PASA dataset.

Furthermore, in the empirical results (Table \ref{tab:model_comparison}), although the performance of TransTab was not the best among the baseline models, it exhibits greater resistance to data corruption, particularly at high corruption rates. This resilience might be due to its incorporation of unstructured free-texts into the modeling process.

These findings suggest that unstructured free-text information within EHRs may play a pivotal role in enabling our model with strong resilience against higher levels of corruption in structured EHRs.

\subsection{Limitations and future work}
There is still room for improvement in our present study. 
Currently, our work involved the design of hard prompts, which consisted of hand-crafted and interpretable text tokens. We may consider further exploration of prompt learning in future work, including the investigation of diverse designs of medical prompts and the utilization of soft prompts with less explicit instructions that can be optimized in training for specific tasks. 
Additionally, the EHR data used in our analysis was acquired solely at a singular time instance, consequently failing to capture longitudinal patient information in EHRs. For future work, our focus will be on incorporating temporal information through prompt learning in EHRs.
Lastly, further external studies on other populations should be performed to validate the feasibility of the framework in clinical practice.

\section{Conclusion}
This study demonstrates that the proposed framework leveraging NLP techniques is an effective and accurate deep learning approach for predicting patient mortality and improving resource utilization in critical care through multimodal EHRs.
It also reveals several key findings. 
Firstly, the predictive performance of our proposed model consistently outperformed the leading baselines. 
Secondly, the experimental results strongly support leveraging prompt learning with a transformer encoder to effectively model multimodal EHRs.
Finally, our proposed framework demonstrated high resilience to data corruption in structured EHRs, indicating its strong feasibility in real clinical settings.

\section*{Declarations}
\subsection*{Ethical Approval and Consent to participate}
MIMIC-III: Not applicable.
PASA: The SingHealth Centralised Institutional Review Board (SingHealth CIRB 2014/651/D, 2020/2915, and 2021/2547) has granted approval to the database with a waiver of patient consent.

\subsection*{Consent for publication}
Not applicable.

\subsubsection*{Availability of supporting data}
The MIMIC-III data used in this study are available in the Physionet Repository, at \href{https://physionet.org/content/mimiciii/1.4/}{https://physionet.org/content/mimiciii/1.4/}. Access to MIMIC dataset is available only to individuals who fulfill the credentialing requirements and agree to the terms of use.
The PASA data underlying this article were provided by Singapore General Hospital by permission, and will be shared on request to the corresponding author with permission of Singapore General Hospital. 

\subsubsection*{Competing interests}
The authors have no competing interests to declare.

\subsubsection*{Funding}
iThis research is supported by the National Research Foundation Singapore under its AI Singapore Programme grant number AISG2-TC-2022-004. This research is also supported by A*STAR, CISCO Systems (USA) Pte. Ltd and National University of Singapore under its Cisco-NUS Accelerated Digital Economy Corporate Laboratory (Award I21001E0002).

\subsubsection*{Authors' contributions}
\textbf{Yucheng Ruan:} Conceptualization, Data curation, Methodology, Software, Investigation, Formal analysis, Visualization, Writing - Original Draft, Writing - Review \& Editing. \textbf{Xiang Lan:} Conceptualization, Data curation, Software, Investigation, Writing - Review \& Editing. \textbf{Daniel J. Tan:} Data curation, Software, Investigation. \textbf{Hairil Rizal Abdullah:} Supervision, Writing - Review \& Editing. \textbf{Mengling Feng:} Conceptualization, Supervision, Funding acquisition, Writing - Review \& Editing.

\subsubsection*{Acknowledgements}
Not applicable.

\bibliography{sn-bibliography}

\end{document}